\documentclass[Afour,sageh,times]{sagej}

\usepackage{moreverb,url}
\usepackage{multicol}
\usepackage{xifthen}

\usepackage[colorlinks,bookmarksopen,bookmarksnumbered,citecolor=red,urlcolor=red]{hyperref}

\usepackage{multicol}
\usepackage{graphics} 
\usepackage{mathptmx} 
\usepackage{amsmath} 
\usepackage{amssymb}  
\usepackage[noend]{algpseudocode}
\usepackage{subfig}
\usepackage{algorithmicx,algorithm}
\usepackage{multirow}
\usepackage{hyperref}
\usepackage{mathrsfs}
\usepackage{bbm}
\usepackage{tabularx}
\usepackage{color}
\usepackage{graphicx}
\usepackage{caption}

\usepackage{threeparttable}
\usepackage{booktabs}
\usepackage{graphicx}
\usepackage{xspace}
\usepackage{float}

\renewcommand{\eqref}[1]{(\ref{#1})}
\newcommand{\figref}[1]{Fig.~\ref{#1}}
\newcommand{\subfig}[1]{\textit{#1}}
\newcommand{\tabref}[1]{Table~\ref{#1}}

\newcommand{\ie}{\textrm{i.e.}}
\newcommand{\eg}{\textrm{e.g.}}

\newcommand{\invigorate}{\mbox{{\small INVIGORATE}}\xspace}

\newcommand{\numobj}{N}
\newcommand{\obj}{i}

\newcommand{\tgt}{i^*}
\newcommand{\ntgt}{i^-}
\newcommand{\objbox}{B}
\newcommand{\objboxset}{\mathcal{B}}
\newcommand{\objboxdet}{\mathcal{B}^D}
\newcommand{\objboxhis}{\mathcal{B}^H}
\newcommand{\objclsscore}{c}
\newcommand{\image}{\ensuremath{I}\xspace}

\usepackage{xcolor}

\newcommand{\expr}{\ensuremath{E}\xspace}
\newcommand{\ans}{ans}
\newcommand{\ansr}{\ans_r}
\newcommand{\ansd}{\ans_d}

\newcommand{\resp}{Res_p}
\newcommand{\resn}{Res_n}
\newcommand{\randomtest}{Test A\xspace}
\newcommand{\hardtest}{Test B\xspace}
\newcommand{\overalltest}{Overall\xspace}

\newcommand{\state}{s}
\newcommand{\gstate}{s^{g}}
\newcommand{\rstate}{s^{r}}
\newcommand{\belief}{b}
\newcommand{\gbelief}{b^{g}}
\newcommand{\rbelief}{b^{r}}
\newcommand{\obs}{o}
\newcommand{\action}{a}
\newcommand{\gaction}{\action^g}
\newcommand{\qaction}{\action^q}

\newcommand{\aspace}{A\xspace}
\newcommand{\rew}{R\xspace}
\newcommand{\tmodel}{T\xspace}
\newcommand{\omodel}{Z\xspace}

\newcommand{\gnet}{G-Net\xspace}
\newcommand{\rnet}{R-Net\xspace}
\newcommand{\onet}{O-Net\xspace}
\newcommand{\cnet}{Q-Net\xspace}

\newcommand{\graspmacro}[1]{Get({}#1)}
\newcommand{\askquestion}[1]{AskQuestion({}#1)}
\newcommand{\hungcost}[2]{H({}#1,{}#2)}
\newcommand{\iou}[2]{\frac{|{}#1\cap{}#2|}{|{}#1\cup{}#2|}}

\newcommand{\gscoresym}{g}

\newcommand{\gnetsym}{G}

\newcommand{\rnetsym}{R}
\newcommand{\rscoresym}{r}

\newcommand{\capset}{\mathcal{Q}}
\newcommand{\qcaption}{q}
\newcommand{\selfcaption}[1]{\qcaption_S^{#1}}
\newcommand{\relcaption}[1]{\ifthenelse{\isempty{#1}}{\qcaption_R}{\qcaption_R^{#1}}}
\newcommand{\mixcaption}[1]{\qcaption_M^{#1}}
\newcommand{\selfcapset}[1]{\ifthenelse{\isempty{#1}}{\capset_S}{\capset_S^{#1}}}
\newcommand{\relcapset}[1]{\ifthenelse{\isempty{#1}}{\capset_R}{\capset_R^{#1}}}
\newcommand{\mixcapset}[1]{\ifthenelse{\isempty{#1}}{\capset_M}{\capset_M^{#1}}}

\newcommand\BibTeX{{\rmfamily B\kern-.05em \textsc{i\kern-.025em b}\kern-.08em
T\kern-.1667em\lower.7ex\hbox{E}\kern-.125emX}}

\usepackage{color}
\definecolor{fullred}{rgb}{0.95,.0,.1}
\newcounter{cmt}

\newcommand{\bluetxt}[1]{#1}

\def\volumeyear{2016}

\begin{document}

\runninghead{Zhang et al.}

\title{INVIGORATE: Interactive Visual Grounding and Grasping in Clutter}

\author{Hanbo Zhang\affilnum{1,2}, Yunfan Lu\affilnum{2}, Cunjun Yu\affilnum{2}, David Hsu\affilnum{2}, Xuguang Lan\affilnum{1} and Nanning Zheng\affilnum{1}}

\affiliation{\affilnum{1}Xi'an Jiaotong University, CN\\
\affilnum{2}National University of Singapore, SG}


\email{zhanghanbo163@stu.xjtu.edu.cn}

\begin{abstract}

This paper presents~\invigorate, a robot system that uses natural language to communicate with humans and grasps a specified object in cluttered environments. 
It addresses several challenges: (i) inferring the target object among other
occluding objects, from input language instructions and RGB images, (ii)
inferring object blocking relationships (OBRs) from the images, and (iii)
synthesizing a multi-step plan to ask disambiguation questions about the
target object and to grasp the object successfully.  \invigorate contains
several neural network modules trained for object detection, for visual
grounding, for question generation, and for OBR detection and grasping. It
allows for unrestricted object categories and language expressions, subject to
the training datasets.
\bluetxt{To overcome \textit{uncertainties} in visual perception and \textit{ambiguity} in languages,
 \invigorate employs an object-centric partially observable Markov decision process (POMDP) that integrates the learned neural network modules.
 It tracks the relevant objects by maintaining a history of observations and asks disambiguation questions when necessary.
 Using the POMDP, it searches for a near-optimal sequence of actions that identify and grasp the target objects.}
\invigorate leverages the advantages of model-based POMDP planning and
data-driven deep learning to address the challenge of  complex, high-dimensional observations and achieve robust robot performance under uncertainty.
 Experiments with \invigorate on a Fetch robot indicate significant benefits
 of this integrated approach.
A demonstration video is available at \href{https://youtu.be/rQdBr\_yeVjA}{https://youtu.be/rQdBr\_yeVjA}. 

\end{abstract}



\maketitle 

\section{Introduction}

\begin{figure}

    \begin{tabular}{cc}
       \includegraphics[height=70pt]{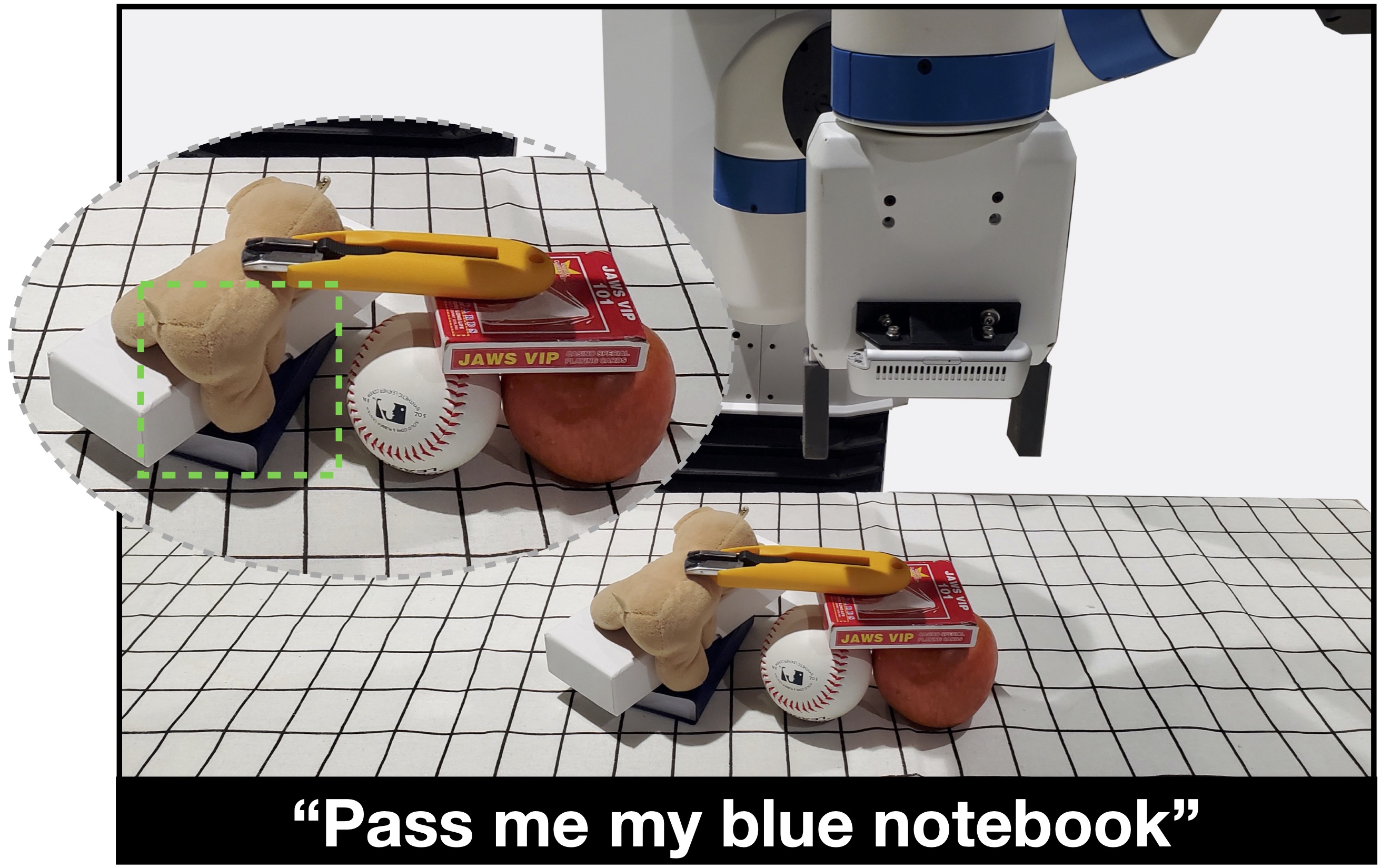}
   &        \includegraphics[height=70pt]{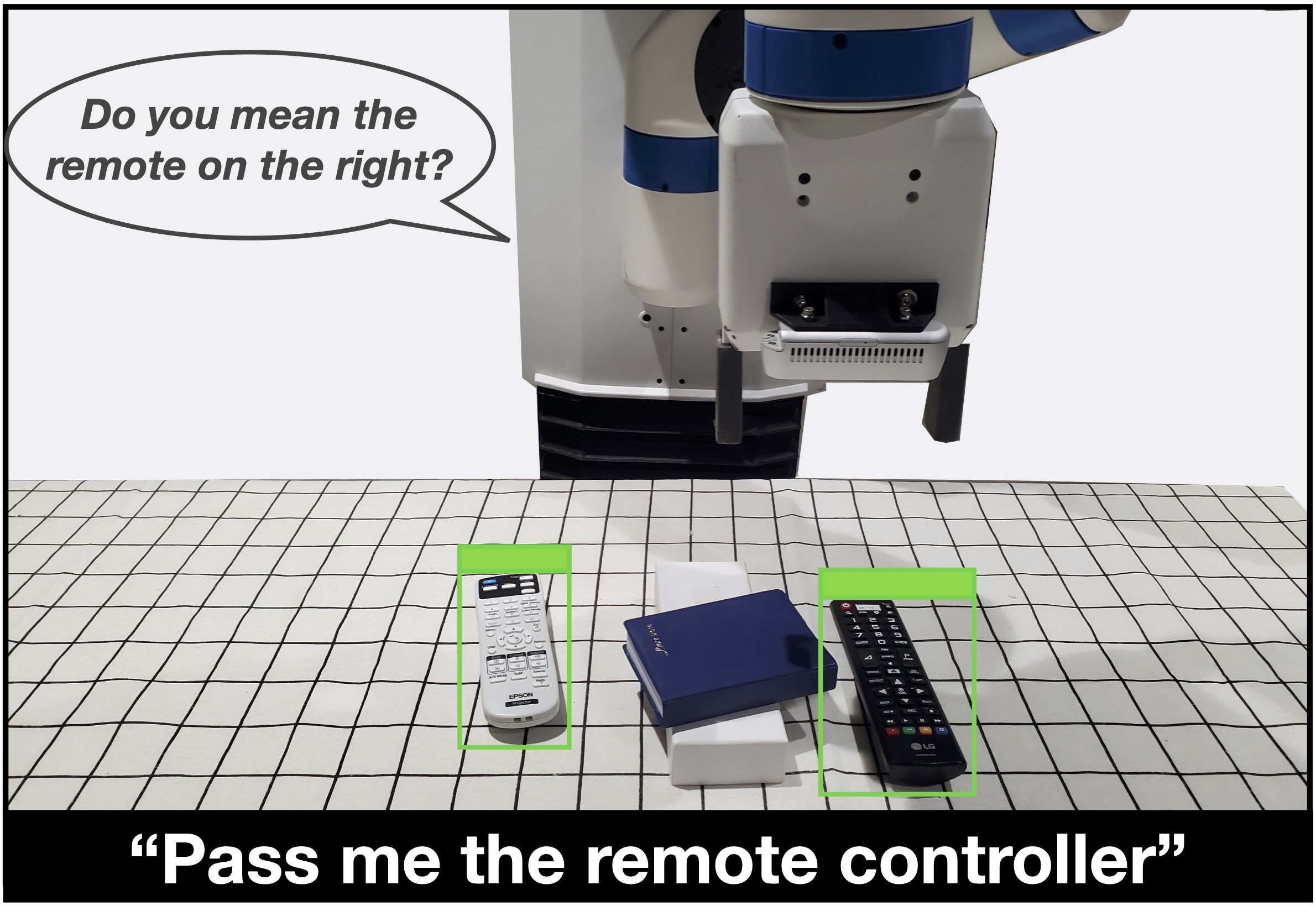}  \\
      (\subfig a)    & (\subfig b) 
    \end{tabular}
    \caption{Interactive visual grounding and grasping in clutter. The robot
      receives a verbal instruction from the human to retrieve an
      object. It tries to identify the target object visually, asks questions
      to disambiguate the target object, if necessary, and eventually grasps
      the object. (\subfig a) Perceptual uncertainties. The object detection
      module fails to detect the notebook because of visual
      occlusion. (\subfig b) Language ambiguity.  The  instruction is
      ambiguous. There are two remote controllers, one black and one white.
      Since both satisfy the instruction, the robot asks questions to disambiguate.}
    \label{fig:motivation}
\end{figure}

Robots are gradually, but surely entering into our daily life. To become
effective human helpers, robots have to understand our physical world through
visual perception and interact with humans through natural  language.
 Consider the robot task of following a human verbal instruction to retrieve an object from a cluttered kitchen table (\figref{fig:motivation}). This seemingly simple task presents multiple challenges:
\begin{itemize}
    \item Infer the target object among other occluding objects from the input language instruction and images;
    \item Infer  object blocking relationships  from images;
    \item Synthesize a multi-step plan to  disambiguate the target object
      by asking questions  and retrieve the target despite other obstructing objects.  
\end{itemize}
Advances in deep learning have yielded powerful neural network (NN) models to process complex visual and language inputs, addressing the first two challenges. However, they alone are not sufficient for two main reasons. 
Firstly, visual inputs are complex and noisy, leading to errors in perceptual processing. 
Cluttered scenes are inherently \textit{partially observable} and exacerbate the difficulty of perceptual processing. For example, the target object may not be detected at all because of visual occlusions (\figref{fig:motivation}\subfig a).  Secondly, despite their richness,  human languages are sometimes ambiguous. Two distinct objects may both match the language specification  (\figref{fig:motivation}\subfig b). 
A natural question then arises: \textit{How can we harness the power of these learned NN models for perceptual and language processing and achieve robust robot performance?} 

To this end, we have developed and experimented with a robot system,
\textit{INteractive VIsual GrOunding and gRAsp in clutTEr} (\invigorate).  See
\figref{fig:motivation} for examples.  \invigorate integrates data-driven
learning and model-based planning. To address complex visual inputs and
language interactions, we train separate NN models to detect objects,
to ground verbal references, to generate questions that
disambiguate object references, and to estimate object spatial relationships
for grasping.  These NN models are then integrated into an object-centric partially observable
Markov decision process (POMDP). 
In the INVIGORATE POMDP, we model the NN outputs---the detected target object, other
objects, and object blocking relationships---as noisy observations and learn a
probabilistic observation model of detection failures.  Through POMDP
planning, \invigorate tracks the history of observations on individual objects
over time and obtains a robust probabilistic estimate of the objects and their
relationships, despite uncertainties in perceptual and language processing.

If a verbal reference to the target object is ambiguous, \invigorate gathers information \emph{actively} by asking  disambiguation
questions. It reasons systematically about the uncertainty
of the target object and balances the potential benefit of additional
information  against the cost of asking questions.
\bluetxt{Furthermore, it  optimizes the generated questions by 
predicting  possible interpretations by humans.}
  
We deployed \invigorate on a Fetch robot. Experimental results show that \invigorate achieves an overall success rate of 83\% on our test dataset and consistently outperforms a baseline without POMDP integration. Ablation studies further confirm the importance of reasoning about uncertainties in dealing with noisy visual perception and language ambiguity.

One main contribution of this work is to demonstrate  a principled approach
that integrates data-driven deep learning and model-based planning for 
complex robot tasks.  We build a POMDP model  connecting three key elements:
robot perception, object manipulation, and human-robot language interaction.
The learned NN models enable \invigorate to handle complex visual inputs and
language interactions. Model-based POMDP planning enables \invigorate to
achieve robust overall performance under  uncertainty in perceptual and
language processing.


\section{Related Work}
\label{sec:related-work}


\subsection{Visual Grounding of Referring Expressions}

\bluetxt{ Visual grounding connects language and vision: it links a language
  expression with its corresponding visual representation in an image. Given a
  language instruction and visual observations of the scene, \invigorate tries
  to grasp a target object in clutter.  As a first step, it solves a visual
  grounding problem: recognize and locate in the input image  the target object specified by the language
  instruction.  Visual grounding has been studied
  extensively in computer vision~\citep{qiao2020referring}, especially, with
  deep learning techniques ~\citep{guadarrama2014open, nagaraja2016modeling,
    yu2018mattnet}.}
\bluetxt{ Earlier work typically formulates visual grounding as matching a
  natural-language referring expression and a set of object proposals from the
  corresponding image, and uses learning to
  optimize the matching score.  Transformers \citep{vaswani2017attention} have
  brought significant new progress \citep{lu2019vilbert, chen2020uniter,
    deng2021transvg, li2021referring, kamath2021mdetr, zeng2022xvlm}, by 
  predict directly  bounding boxes of referred objects. These methods,
  however, implicitly assume that objects are clearly visible with little occlusion.
}

\bluetxt{
  To guard against occlusion and other perceptual uncertainties, \invigorate integrates
  observations over time and interacts with human
  actively via natural language for disambiguation,
  leading to an effective and robust system.
}

\subsection{Human-Robot Interaction  via Natural Language }

\bluetxt{
To resolve ambiguity in language instructions, the robot may ask the human
clarification questions.  
Some earlier robot systems have made important advances: they interact with
humans verbally to seek help, when facing ambiguities during navigation
\citep{kruijff2006clarification, hemachandra2015information}, when collaborating
with humans \citep{rosenthal2010effective}, or when encountering failures
\citep{deits2013clarifying, tellex2014asking}.
However, these systems are typically deployed in restrictive settings, as a result of limited model capacity and training data size.
}

\bluetxt{ Recent image captioning models based on deep
  learning~\citep{bernardi2016automatic, hossain2019comprehensive,
    katiyar2021comparative} show clear advantages over traditional
  approaches. They take raw image features as input and directly generate
  captions end-to-end in an auto-regressive manner \citep{hochreiter1997long,
    vaswani2017attention}.  Dense captioning models, in particular, generate
  captions for objects individually \citep{johnson2016densecap,
    yang2017dense}. After training on large-scale datasets, they produce much
  richer and more natural object descriptions.
  INGRESS~\citep{shridhar2020ingress} leverages these advances to generate both
  entity and relational expressions for a target object. 
  The generated expressions are then fitted into a question template to generate
  disambiguation questions.
  Again, visual occlusion may degrade the performance of these
  models, resulting in noisy or even completely wrong expressions.}

\bluetxt{
To address the difficulty, INVIGORATE first generates a set of
candidate questions. It predicts how humans interpret these questions
and selects the best question through POMDP planning
 to optimize the overall performance over time. 
}

\subsection{Goal-directed Object Grasping in Clutter}

Object grasping is widely studied in robotics, with a vast
literature~\citep{bohg2013data}. In recent years, both data-driven learning
and model-based planning have contributed to significant progress in object
grasping in general~\citep{lenz2015deep, redmon2015real, mahler2017dex,
  mahler2019learning, garg2019learning} and goal-directed object grasping in
particular~\citep{jang2017end, fang2018multi, zengrobotic, zhang2019multi,
  murali20206}. It is beyond the scope of this paper to provide a
comprehensive survey. We give only a few selected examples closely related
to our task.

Several recent methods aim at object retrieval in cluttered scenes
\bluetxt{
  \citep{guo2016object, zeng2018learning, danielczuk2019mechanical,
    kurenkov2020visuomotor, yang2020deep}.}
In particular, \citet{zhang2019multi} propose to learn object-blocking
relationships for grasping.  However,  these earlier methods do not address the issue of language
interaction between the human and the robot.  \citet{hatori2018interactively}
and~\citet{chen2021joint} propose to fuse visual and text features in neural
networks \bluetxt{and grasp the target object based on natural language
  instructions.}  \citet{shridhar2020ingress} formulate a POMDP to ask
disambiguation questions about the target object in interactive grasping
tasks.  \citet{mees2021composing} propose a robot system capable of grounding
language instructions for both object picking and placement.  They, however, do not
consider visual occlusion and physical obstruction in dense object clutter.
\bluetxt{
  \invigorate aims to tackle the dual challenges of object grasping in clutter
  and human-robot natural language interaction together.}


\subsection{Integration of Learning and Planning}

One key strength of \invigorate  is  the integration of learning and planning,
an active research direction that has attracted much attention recently.

Learning and planning interact in various ways.  \bluetxt{One most common
  approach is to learn models for planning.  Models of environment dynamics,
  observations, and task reward are critical for planning, but are notoriously
  difficult to design manually, especially for complex systems. One approach is
  to learn these models from data, through supervised
  model learning, model-based reinforcement learning \citep{moerland2023model},  or inverse
  reinforcement learning \citep{arora2021survey}.
We may even do so in    the latent space, through representation
learning \citep{hafner2019learning,sekar2020planning, okada2021dreaming}.
}


  \bluetxt{One recent, highly successful idea is to insert learned
    policies or value functions into planning algorithms \citep{silver2016mastering, cai2019lets}.
The learned policies and value functions dramtically speed up forward-search
planning algorithms by providing an effective search heuristic and reducing
the search horizon.}

\bluetxt{
  An opposite, but equally interesting idea is to embed a planning algorithm,
  or more broadly, any robot algorithm, 
  into a neural network as the structure prior for end-to-end learning
  \citep{tamar2016value}.  These \textit{Differentiable Algorithm
    Networks}  (DANs) \citep{karkus2019differentiable}
cover a wide variety of essential robot algorithms for state
estimation~\citep{jonschkowski2018differentiable, karkus2018particle},
planning under full or partial observability
\citep{tamar2016value,karkus2017qmdp,guez2018learning}, and control
\citep{amos2018differentiable} .   They are structured, interpretable,
  task-driven, and robust by combining the benefits of model-based planning
  and data-driven learning.}


INVIGORATE does not fall into any of the categories above. It builds on top of
a set of NN models learned from data for vision and language processing; it
then applies a model-based approach to integrate these NN modules and reasons
about their uncertainties systematically in order to achieve robust robot
performance.

\bluetxt{ This paper extends our earlier work \citep{zhanglu2021invigorate}.
  We introduce a novel question generation method for disambiguation.  We also provide additional
  experiments and analyses. }


\section{Overview}


\begin{figure}
 \center{\includegraphics[width=0.48\textwidth]{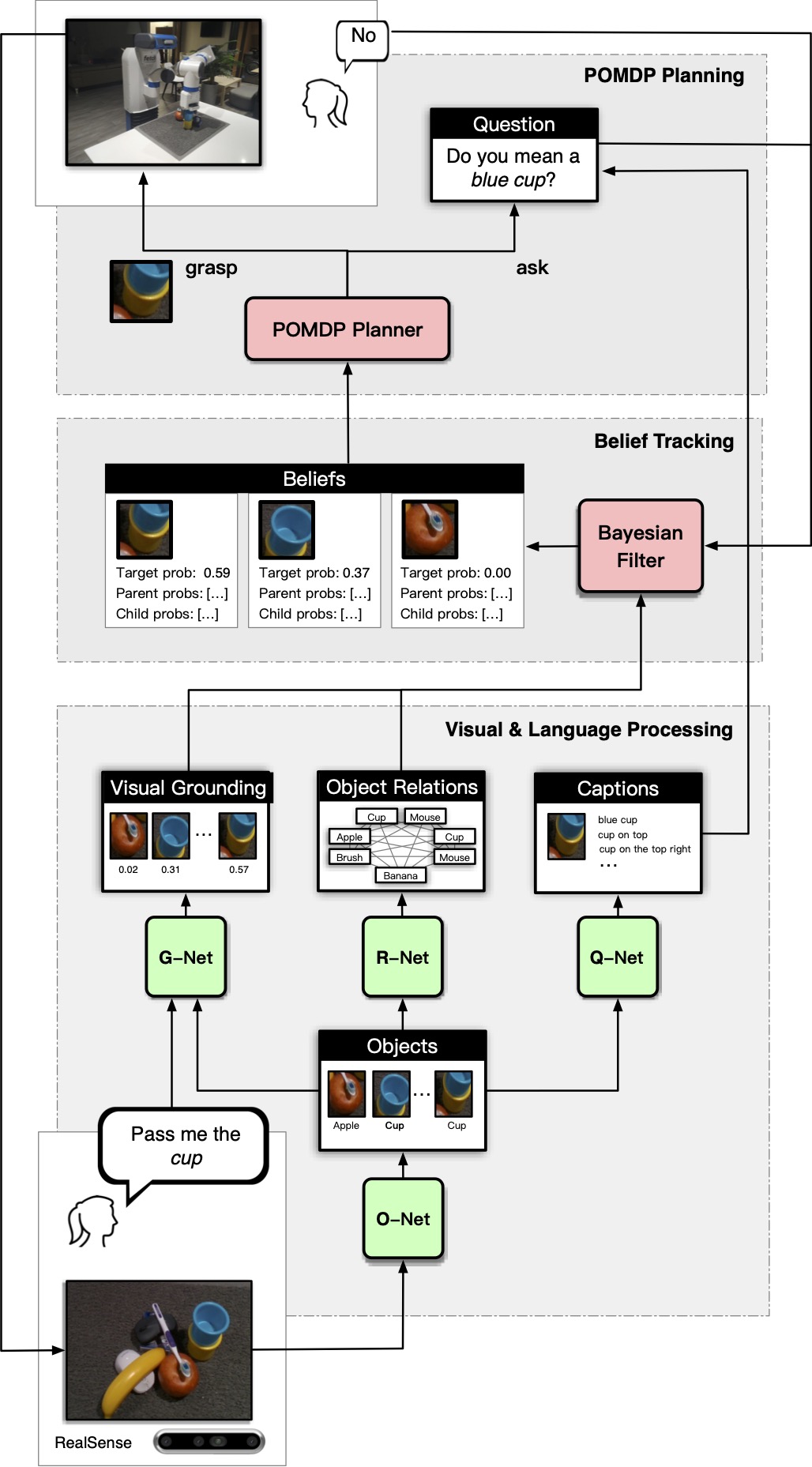}}
 \caption{An overview of INVIGORATE. 
 \bluetxt{INVIGORATE integrates data-driven deep learning with model-based POMDP planning. It consists of three components: POMDP planning (top), belief tracking (middle), and visual and language processing (bottom). 
}
}
 \label{fig:overview}
 \end{figure}

 \invigorate takes a verbal instruction from the human to grasp an object of
 interest in clutter. It uses both the natural language instruction and images from its visual
 sensor to identify the target object. If the referred object in
 the instruction is ambiguous, \invigorate asks the human simple questions for
 disambiguation and eventually grasps the target object, while  avoiding unnecessarily
 perturbing other objects. See \figref{fig:overview} for an overview.

 \invigorate integrates data-driven deep learning with model-based POMDP
 planning.  We train four NN modules, \onet, \rnet, \gnet, and \cnet, from
 data for visual and language processing (\figref{fig:overview}, bottom). At each time step, \onet generates from the input image $\image$ a set
 of object proposals. Based on these object proposals, \rnet further processes
 \image and extracts pairwise blocking relationships among the objects, as
 well as candidate object grasps. \gnet uses the referring expression
 \expr in the verbal instruction and the object proposals for visual grounding; it outputs a set of
 candidates for the target object. If \expr is ambiguous, there may be
 multiple candidates. \invigorate uses \cnet to generate a set of referring
 expressions of all candidates. It fits the generated expression into a
 question template and asks the human a disambiguation question, \eg, ``Do you
 mean the cup on top?''.

 The trained NN models are powerful and allow for unrestricted object
 categories and language expressions, subject to the training
 datasets. However, the outputs of \onet, \rnet, and \gnet are all noisy,
 because of sensor noise, visual occlusion, and ambiguity in human
 languages. To achieve robust robot performance under these uncertainties,
 we build a POMDP model that integrates the learned NN modules.  \invigorate
 maintains a \textit{belief}, \ie, a  probability distribution over the
 underlying state.
 \invigorate uses a factored, object-centric state representation that
consists of the target object, the other objects, and their blocking
relationships.
 It treats the NN outputs as \textit{noisy  observations}  on the objects and
 their relationships, in order to
 account for the NN prediction errors.  At each step, \invigorate updates the
belief with new observations and actions, through
 Bayesian filtering (\figref{fig:overview}, middle).  The belief summarizes the history of observations and
 actions; it quantifies the uncertainties probabilistically and provides the
 basis for a principled approach to robust robot decision making.

 Given a belief, \invigorate performs approximate POMDP planning through
 look-ahead search to choose the best action (\figref{fig:overview}, top). \invigorate models two types of
 actions: ask a disambiguation question or grasp a target object. Intuitively,
 if the belief over the target object has high uncertainty, 
 \invigorate \bluetxt{may ask a disambiguation question} to
 gain additional information.  \bluetxt{Further, it selects the question
   carefully  by predicting possible human interpretations, thus avoiding uninformative
   questions and enabling natural human interaction.}
 If the belief has low uncertainty, \invigorate grasps either the
 target object directly or an obstructing object according to the estimated
 object blocking relationships.  By reasoning about the belief, POMDP planning
 enables \invigorate to choose a near-optimal sequence of actions.


\section{Neural Networks for Visual Perception and Language Interaction}


\invigorate uses deep learning to build perceptual and interaction modules. We describe each of these modules in the subsections below.
\subsection{\onet for Object Detection and Tracking}


\invigorate applies the Cascade R-CNN~\citep{cai2018cascade} as the
base object detector, which is trained on the union of COCO~\citep{lin2014microsoft} and
VMRD~\citep{zhang2018visual}.
\bluetxt{
Nevertheless, imperfect detection is inevitable due to noisy visual observations, neural network failures, and partial observability in our tasks.
Moreover, \invigorate is trying to deal with multi-step goal-directed grasping, which means that it needs to track the object states and maintain consistency throughout multi-step results.
Hence, we introduce an object detection and cross-step tracking mechanism in this section. 
}

Specifically, to track objects across multiple steps, we maintain an object pool $\objboxset$.
In each step, we first feed the raw image \image to the base detector and obtain a set of proposals $\objboxdet$.
Then, we feed all historical proposals in $\objboxset$ to the object detector to re-classify them and get a historical set $\objboxhis$. 
Subsequently, we merge $\objboxhis$ into $\objboxdet$ 
using Hungarian algorithm with the cost function defined as:
\begin{align} \label{eq:hungcost}
\hungcost{\objbox_\obj}{\objbox_j}=\alpha_1\iou{\objbox_\obj}{\objbox_j} + \alpha_2\left\| \objclsscore_\obj - \objclsscore_j \right\|
\end{align}
\bluetxt{where $\objbox_\obj$ and $\objbox_j$ represent the bounding boxes of object $\obj$ and $j$, $\iou{\objbox_\obj}{\objbox_j}$ is the \textit{intersection of union} (IoU) between $\objbox_\obj$ and $\objbox_j$, $\objclsscore_\obj$ and $\objclsscore_j$ mean the normalized confidence scores given by the object detector, and $\alpha_1$ and $\alpha_2$ are the weights to balance the two items and $\alpha_1 +\alpha_2 = 1$.}
Intuitively, bounding boxes with large IoU and the same category will be merged into one.
Finally, the object pool $\objboxset$ will be updated by $\objboxdet\cup\objboxhis$ using Hungarian algorithm again with the same cost function defined in \eqref{eq:hungcost}.
Such a detection procedure is more robust against false positives and false negatives.
The merging process based on the Hungarian algorithm also enables object tracking across different steps, which is a prerequisite for belief updates.

\subsection{\gnet for Visual Grounding}

\gnet takes an image $\image$, a referring expression $\expr$, and detected object proposals in $\objboxset$ to estimate the matching scores between each detected object $\obj$ and the referring expression $\expr$:
\begin{align}
\gscoresym_\obj=\gnetsym(\objbox_i,\expr,\image)
\end{align}
where $\gnetsym$ denotes \gnet and $\gscoresym_\obj$ denotes the output matching score.
In \invigorate, we train \gnet following \citet{yu2018mattnet} on the RefCOCO dataset.
\gnet splits the user expression into three parts: the subject description, the locational description, and the relational description. 
It extracts the visual feature for each proposal and performs separate visual-text matching. To illustrate, given the expression ``the blue cup to the right of the book'', such a sentence would be decomposed into a subject description ``the blue cup'', a locational description ``to the right of'' and a relational description ``the book''. An embedding is obtained for each description through a language attention network. 
The image is fed into a CNN-based neural network to generate a visual feature map, which is then pooled into region features using proposals in $\objboxset$.
{After that, we estimate a similarity score between phrases and pooled visual features together with their location information with a trainable layer. 
Finally, an attention-based summation of similarity scores for three phrases is used to determine the referred object.
All layers are trained in an end-to-end manner.}
As a result, \gnet is scalable and can handle almost unrestricted referring expressions. \eg, ``the red apple'', ``the apple on top'', ``the blue cup to the right of the book'', etc. 
\subsection{\cnet for Question Generation}
\label{sec:qgen}
To ask  a disambiguation question, \cnet first generates a
caption about an image region and then fits it into a question template.
\bluetxt{ Specifically, it uses 
  two captioners: a \textit{self-referential} captioner  for
  object attributes (\eg, object category, color, or shape)
  and a \textit{relational} captioner for
  object relationships.
However, visual occlusions can cause noisy and sometimes completely misleading captions, particularly for the self-referential ones.
 To address this issue, \invigorate takes \gnet as a human response model and selects the best among the generated candidates. 
}

\bluetxt{ \cnet uses DenseCap \citep{johnson2016densecap} as the self-referential captioner.
It takes as inputs object features derived from its bounding box and then generates descriptions using LSTM \citep{hochreiter1997long} in an auto-regressive manner, i.e., words are generated one by one based on the input features and all previous words.
For each object $\obj$, it generates a noisy object-specific caption $\selfcaption{\obj}$, and thus, it constructs a self-referential caption set $\selfcapset{}=\{\selfcaption{\obj}\}_{\obj=1}^{\numobj}$ for the workspace, where $\numobj$ is the number of detected objects.}

  \cnet uses  UMD RefExp \citep{nagaraja2016modeling, shridhar2020ingress}
  as the relational captioner.  
It takes as input the visual and spatial features of the object of interest $\obj$ and a context object $i_c$, which may be the whole image.
The generation of relational captions is similar to self-referential captions, also based on auto-regression and LSTM.
To ensure that \bluetxt{the relational description of the object $\obj$} is informative, we feed all possible pairs $(\obj, i_c)$ consisting of object $\obj$ and every possible context object $i_c$ to the relational captioner, and choose the one with maximum confidence score:
\begin{align}
\relcaption{\obj}=\arg\max_{\relcaption{}} P(\relcaption{}|\obj, i_c, \image)
\end{align}
\bluetxt{Relational captioner finally outputs a set of relational captions $\relcapset{}=\{\relcaption{\obj}\}_{\obj=1}^{\numobj}$ for every object $\obj$.}

\bluetxt{In addition, \cnet generates a set of mixed captions
  $\mixcapset{}=\{\mixcaption{\obj}\}_{\obj=1}^{\numobj}$,
containing both self-referential captions and relational captions to
provide richer, more specific object descriptions.
To  generate a mixed caption for object $i$, we first extract attributes from
  $\selfcaption{i}$ and the relational description from $\relcaption{i}$.
We then concatenate the self-referential
  attributes, the class name, and the relational description into a single expression.}

\bluetxt{
Finally, to choose the best disambiguation question, we use \gnet again as a human response model to predict how humans interpret each caption. We call this \emph{critiqued question generation}. 
Specifically, we choose among all candidate captions
$\capset=\selfcapset{}\cup\relcapset{}\cup\mixcapset{}$ the one that
best matches the interested object according to \gnet:
\begin{align}
    \qaction_i=\arg\max_{\qcaption\in\capset}G(\objbox_i, q, \image) 
\end{align}
\cnet and \gnet play the role of an actor and a critic, respectively.
Experiments show that critiqued question generation provides much more robust
performance, especially, under visual occlusion in clutter.
}

\bluetxt{We follow~\cite{johnson2016densecap} to train our self-referential captioner on Visual Genome dataset \citep{krishna2017visual} and} ~\citet{shridhar2020ingress} to train our relational captioner on RefCOCO dataset \citep{kazemzadeh2014referitgame} using Multi-Instance Learning \citep{foulds2010review}.
Subject to the dataset, our captioners typically generates descriptions such as ``the red apple'' (self-referential), ``the apple on the right of the cup'' (relational) and ``the apple in the back of the image'' (relational).

\subsection{\rnet for OBR and Grasp Detection}
In \invigorate, a single network \rnet outputs both grasps and OBRs of detected objects in $\objboxset$.

For OBR detection, we formulate it as a classification problem, which takes object pairs as inputs and classifies pair-wise OBRs.
{Following \cite{zhang2018visual}, there are three kinds of OBRs: ``parent'', ``child'', and ``none''. 
``Parent'' relation between A and B means A should be grasped after B, and vice versa for the ``child'' relation.}
To classify OBR, we first represent each object by a pooled feature with a fixed size ($7\times7$)\bluetxt{, which is extracted using ROI pooling based on bounding boxes and image features.}
Then we form all possible pair-wise permutations of object features.
The feature of an object pair $(i, j)$ includes the features of $i$, $j$, and the union bounding box.
Finally, the pair-wise OBR for object $(i,j)$ is directly classified based on the corresponding pair-wise feature and results in an OBR score $\rscoresym_{ij}$:
\bluetxt{
\begin{align}
    \rscoresym_{ij} = \rnetsym(\objbox_i, \objbox_j, \image)
\end{align}
where $\rnetsym$ represents the \rnet.}
For grasp detection, since our task is goal-directed, the grasp should be object-specific.
To do so, we detect grasps on each object instead of the input scene.
Concretely, the grasp detector regresses grasps using the $7\times7$ pooled feature of each object with a few convolutional layers.

We follow~\citet{zhang2019multi} to train our \rnet on VMRD~\citep{zhang2018visual}, which contains around 4300 images and 100k grasps.
In practice, we found that the grasp detector sometimes returns unstable grasps.
Therefore, based on the detection result, we finetune the grasp pose through local search.
In detail, we do a grid search by discretizing the area along five dimensions $(x,y,z,w,\theta)$ near the detected grasp, where $(x,y,z)$ is the center of the grasp, $w$ is the width of the gripper, and $\theta$ represents the rotation angle w.r.t. the approaching vector.
We traverse all possible grasp poses to find the best one, whose closing area contains more points of the object.


\section{INVIGORATE POMDP}
\label{sec:invigorate}
\begin{figure}
\vspace{-8pt}
 \center{\includegraphics[width=0.9\columnwidth]{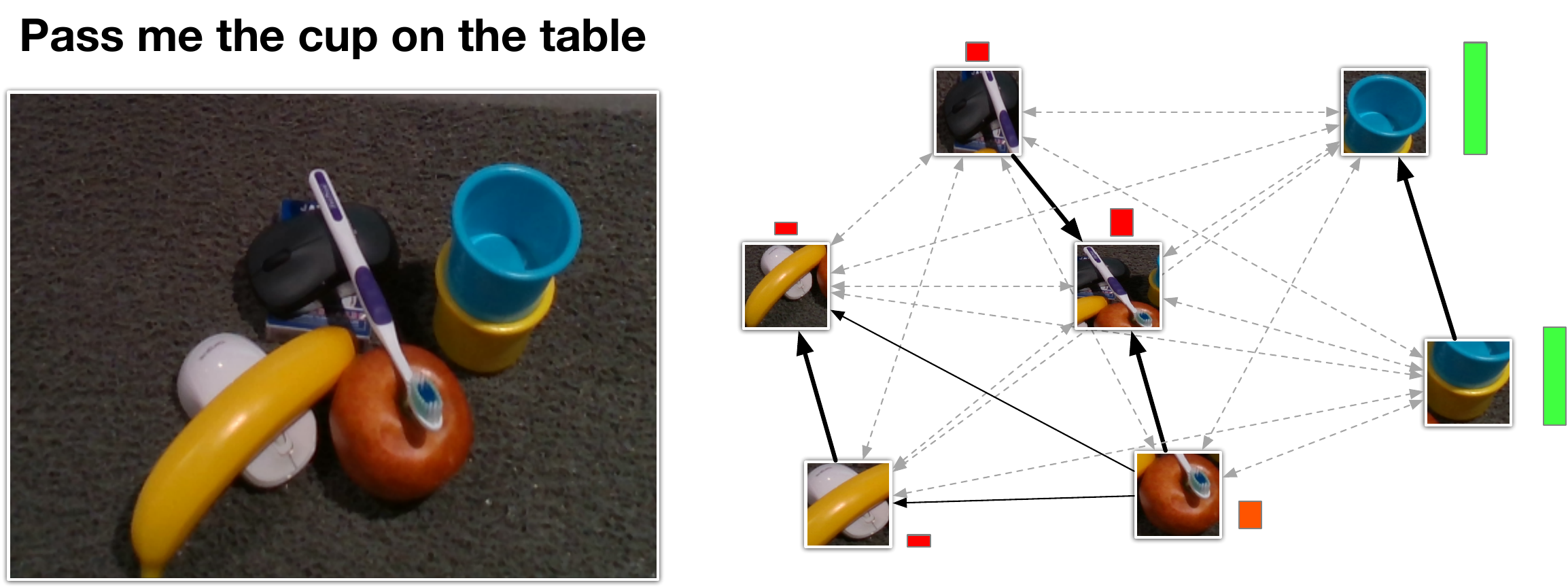}}
 \caption{An example of object-centric belief. Objects are represented using nodes, and the $\gbelief$ is denoted using histograms beside each object. Arrows between objects represent $\rbelief$, meaning that all object relationships are probabilistic. Dashed arrows mean relationships with lower probability.}
 \label{fig:object-centric-belief}
 \end{figure}
 



 To tackle uncertainties in visual perception and ambiguity in language
 intreaction, \invigorate forms a POMDP to integrate the NN modules.

\subsection{State Space}

To grasp the specified target in clutter, the state of \invigorate can be decomposed into two parts, the visual grounding state $\gstate$ and OBR state $\rstate$, i.e., $\state=\gstate \cup \rstate$. 
$\gstate=\cup_{i=1}^{\numobj}\gstate_i$ is an object-centric state~\citep{diuk2008object, wandzel2019multi}, with each $\gstate_i$ indicating whether object $i$ is a target.
$\rstate=\cup_{ij=1}^{\numobj}\rstate_{ij}$ is a graph of all pair-wise OBRs, i.e., the correct grasping order of detected objects, with each $\rstate_{ij}$ denoting the true OBR between object $i$ and $j$.
Since the underlying true state is not available, we maintain a belief $\belief$ at each time step over the state $\state$, which represents a distribution over the state space.
Similarly, $\belief=\gbelief\cup\rbelief$.

We demonstrate an example of object-centric belief in \figref{fig:object-centric-belief}.
Intuitively, the object-centric belief $\belief=\gbelief\cup\rbelief$ in our model is a semantically probabilistic representation of the scene according to visual observations.
It is a noisy estimation of the underlying true state, which will be used for the POMDP decision making.

\subsection{Action Space and Transition Model}
To handle possible ambiguity, \invigorate allows active interaction with human to gather more information.
Therefore,~\invigorate has two types of actions: 1) grasping; 2) asking a question.

\begin{figure}
 \center{\includegraphics[width=\columnwidth]{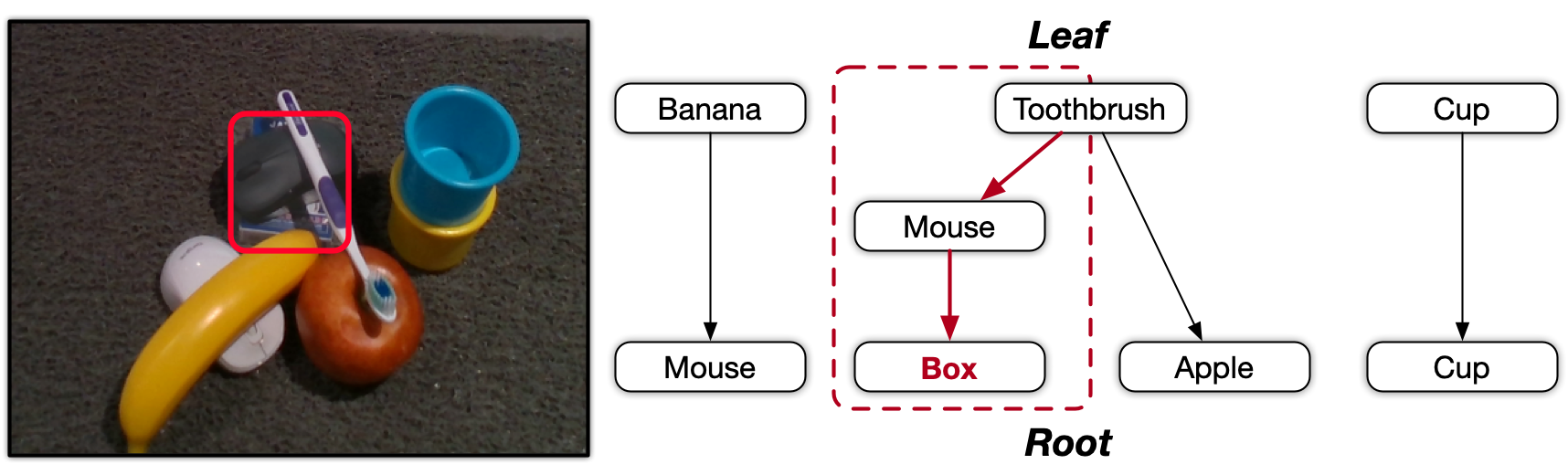}}
 \caption{An example of grasping macro for the blue box in the scene. \textbf{Left}: the scene image with a red box of the target blue box. \textbf{Right}: the object blocking graph of the left image, with a grasping macro marked by red dashed line to grasp the blue box.}
 \label{fig:graspmacro}
 \end{figure}
As shown in \figref{fig:graspmacro}, grasp actions are defined by grasp macros.
Each grasp macro is a sequence of grasps resulting in a terminal state.
Assuming that $\numobj$ objects are detected, there will be $\numobj+1$ grasp macros, including $\numobj$ goal-directed choices and 1 clearing choice.
Each goal-directed grasp macro $\gaction_i$ targets at object $i$. According to $\rbelief$, it sequentially removes exposed objects that most likely blocks object $i$, until it retrieves $i$.
The clearing grasp macro $\gaction_{-1}$ is used to remove all detected objects when none of them is the target.
According to $\rbelief$, it sequentially removes the most exposed objects.
It is non-trivial to analytically find the most exposed object blocking one specified target since all relations are probabilistic.
Therefore, we apply Monte Carlo method to estimate the probability of each object to be exposed and block the target, and then select the most probable one.
Note that in practice, for each step, we only execute the first grasp and then do re-planning, which helps to improve robustness.

For each object $\obj$, action $\qaction_i$ means asking the human whether $\obj$ is the target.
\bluetxt{
The questions are generated from \cnet.
Each question follows a template ``Do you mean $X$?'', where $X$ is a caption from \cnet.
Noticeably, each asking action is associated with a specific object.
To achieve so, we utilize multi-modal interaction \citep{goodrich2008human} by introducing pointing actions when asking questions, i.e., when the robot is asking a question about object $i$, it will also point to it using its end effector.
Therefore, the number of available asking actions is equal to $|\numobj|$.
}

As a result, the size of action space $|\aspace|= 2\numobj + 1$.
Since we assume that human does not change their mind about the target object, for any $\qaction$, the transition model is:
\begin{align}
\tmodel(\state'|\state, \qaction)=\left\{
\begin{aligned}
1,\state'=\state \\
0,\state'\neq\state
\end{aligned}
\right.
\end{align}
On the other hand, any grasp macro results in a terminal state. Therefore, we simply ignore the associated transition model.

\subsection{Visual Observations}
\invigorate takes the output of the \gnet and \rnet as the visual observations after each grasping action.
At time step $t$, we denote the visual grounding observation from \gnet as $\obs^g_t$ and OBR observation from \rnet as $\obs^r_t$.
For brevity, we denote observation $\obs_t$ as $\obs$.
Both $\obs^g$ and $\obs^r$ are object-centric and accord with the state, i.e., $\obs^g=\cup_{i=1}^{\numobj}\gscoresym_{i}$ and $\obs^r=\cup_{ij=1}^{\numobj}\rscoresym_{ij}$.
Accordingly, our visual observation model captures the distribution over $\obs^g$ and $\obs^r$ using visual grounding observation model $\omodel^g$ and OBR observation model $\omodel^r$ in a factorized way.
Formally:
\begin{align}
    \omodel^g=\omodel(\gscoresym_i|\gstate_i)\ \ \ \ \ \ 
    \omodel^r=\omodel(\rscoresym_{ij}|\rstate_{ij})
\end{align}
where $\gscoresym_i$ is the output of \gnet and $\rscoresym_{ij}$ is the output of \rnet. 

Unfortunately, $\omodel^g$ and $\omodel^r$ cannot be specified manually. 
Thus, we resort to data-driven methods.
Specifically, we collect a dataset in clutter
using G-Net, in which each data is represented by $\{\gscoresym_i,\gstate_i\}$ where $\gstate_i$ is a binary label that indicates whether the object $\obj$ is the referred target.
Similarly, we collect a dataset in clutter using \rnet containing tuples $\{\rscoresym_{ij},\rstate_{ij}\}$, where $\rstate_{ij}$ is the ground truth OBR between object $i$ and $j$. We then apply Gaussian kernel density estimation to learn an approximate model for $\omodel^g$ and $\omodel^r$.

\subsection{Textual Observations}

\label{sec:lingobs}

After the robot asks a question $\qaction_t$, it receives an answer from the human, which is the textual observation \bluetxt{$\obs^l_t=\ans$}.
Each textual observation is an unrestricted natural language expression including a response phrase $\ansr$ (e.g. ``Yes" or ``No") that may be followed by an additional description $\ansd$ (e.g. ``No, the left one").
\bluetxt{We assume that the textual observation model is orthogonal to the OBR state $\rstate$.}
To reduce the computation cost of POMDP planning, during the forward search,
we use a simplified observation model over $\ansr$ that effectively ignores the additional description:
\begin{align}
\omodel^l&=\omodel(\ans|\gstate,\qaction)\approx \omodel(\ansr|\gstate,\qaction)
\label{eq5}
\end{align}
where $\ansr$ belongs to either positive phrases $\resp=$\{``Yes'', ``Yeah'', ``Yep'', ``Sure''\} or negative phrases $\resn=$\{``No'', ``Nope''\}.
During the belief update, we handle $\ansr$ according to \eqref{eq5}, but merge $\ansd$ into $\expr$ which will be used for visual grounding in subsequent steps. 

We assume that the human is truthful. Once the human confirms a target, the robot needs not consider other objects anymore.
Under this assumption, the factorized observation model for asking questions is shown in \tabref{table:obsquestion}.
$\epsilon$ is a small positive constant, and in practice, it is set to 0.01.
Intuitively, a positive answer for object $i$ makes object $i$ the only target while a negative answer eliminates object $i$ as a target but does not affect the belief of other objects.

\begin{table}[t] 
\caption{Textual Observation Model $\omodel^l(\ansr|\gstate_i,\qaction)$}
\vspace{-5pt}
\label{table:obsquestion}
\begin{center}
\begin{tabular}{lcc}
\toprule
 & $P(\ansr \in Res_p) $ & $P(\ansr \in Res_n) $\\
\midrule
$\gstate_i=1, \qaction=\qaction_i$ & 1 & 0 \\
\specialrule{0em}{1pt}{1pt}
$\gstate_i=0, \qaction=\qaction_i$ & 0 & 1 \\
\specialrule{0em}{1pt}{1pt}
$\gstate_i=1, \qaction\neq\qaction_i$ & 0 & 1 \\
\specialrule{0em}{1pt}{1pt}
$\gstate_i=0, \qaction\neq\qaction_i$ & $\epsilon$ & $1-\epsilon$ \\
\bottomrule
\end{tabular}
\end{center}
\vspace{-15pt}
\end{table}

\subsection{Reward} 
We want the robot to grasp the correct target while asking a minimal number of questions.
Thus, we impose a small penalty, i.e. a reward of -2, when it asks a question, and a large penalty when it fails the task (e.g. grasping the wrong object).
When multiple objects seemingly satisfy the user expression, the robot cannot accurately differentiate between ambiguity and multi-target.
Thus, to encourage disambiguation and avoid grasping wrong targets in such cases, we empirically engineer the reward for goal-directed grasp macros $\rew(\state,\gaction_i)$:
\begin{align}
\rew(\state,\gaction_i)=\left\{
\begin{aligned} 
\label{eq:graspreward}
&-10+\frac{10}{\sum\gstate}, \ \gstate_i=1 \\
&-10,\quad \quad \quad \quad\gstate_i=0
\end{aligned}
\right.
\end{align}
If there is only one object satisfying the human's instruction, grasping it will result in no penalty.
Otherwise, to encourage disambiguation, the reward of grasping decreases as the number of targets increases.
The robot receives a reward of -10 if it fails to grasp the target.

For the clearing grasp macro $\gaction_{-1}$, the reward is:

\begin{align}
\rew(\state,\gaction_{-1})=\left\{
\begin{aligned} 
\label{eq:clearreward}
&\ 0, \ \ \ \ \ \ \forall \gstate_i=0 \\
&-10, \ otherwise
\end{aligned}
\right.
\end{align}

That is, the robot will not be penalized only if all detected objects are not the target.
Otherwise, it receives a reward of -10 since it removes the target without passing it to the human.

\subsection{Belief Tracking}
\label{sec:belieftrack}

Based on the imperfect observation $\obs$ in each step, we update the belief $\belief_t$ to obtain a more accurate estimate of the underlying true state.
Since our state is object-centric, it can naturally be factorized. 
We factor target belief $\gbelief_{t}$ into belief over each object $\belief_{t}(\gstate_{i})$, and relationship belief $\rbelief_{t}$ into belief over each pair of object $\belief_{t}(\rstate_{ij})$. This factorization allows us to perform belief tracking on each object and object pair separately. 

The robot receives visual observations $\obs^g$ and $\obs^r$ after it performs a grasping and a textual observation $\obs^l$ after it asks a question.
As mentioned, we factor each $\obs^g$ into target observation over each objects, $\obs^g=\cup_{i=1}^{\numobj}\gscoresym_{i}$ and $\obs^r$ into relationship observation over each pair of objects, $\obs^r=\cup_{ij=1}^{\numobj}\rscoresym_{ij}$. 
We then track each factorized belief using Bayesian filter:
\begin{align}
\belief_{t+1}(\gstate_{i})&\propto\omodel^g\cdot \belief_t(\gstate_{i})\nonumber\\
\belief_{t+1}(\rstate_{ij})&\propto\omodel^r\cdot\belief_t(\rstate_{ij})
\end{align}
where $\omodel^g$ and $\omodel^r$ are learned observation model for target and relation respectively. 
Since the human's answer does not affect OBR, $\ans$ is only used to update $\gbelief$\: 
\begin{align}
\belief_{t+1}(\gstate_{i})\propto\omodel^l\cdot\belief_t(\gstate_{i})
\end{align}

\subsection{POMDP Planning}

\begin{figure}
 \center{\includegraphics[width=0.895\columnwidth]{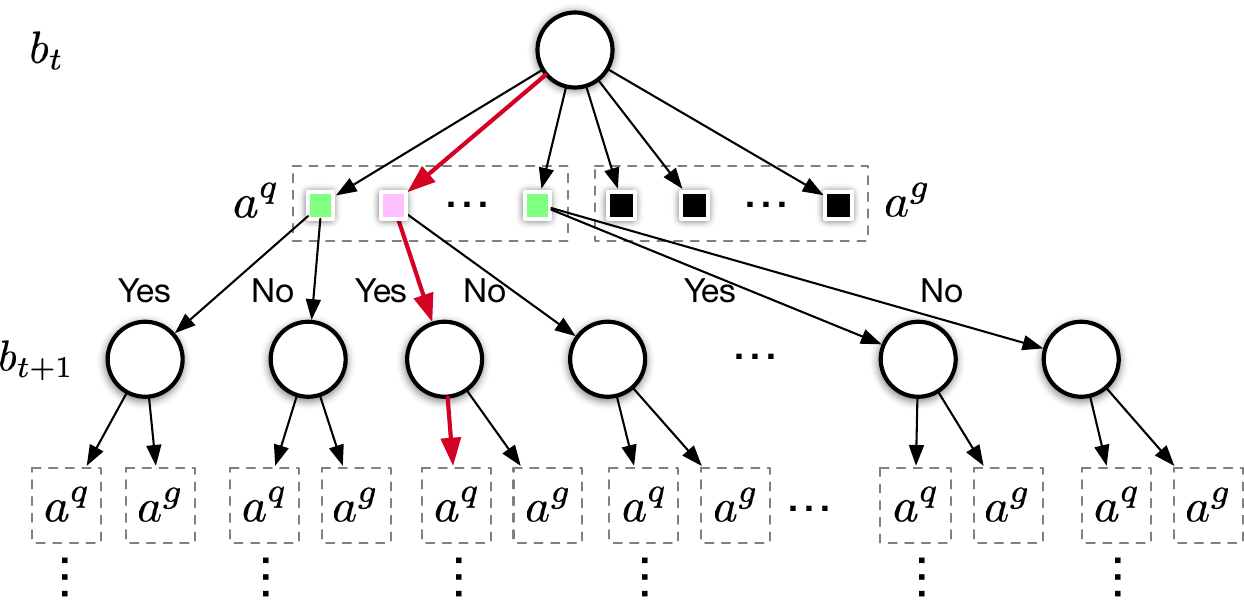}}
 \caption{An overview of policy tree search. Circles denote beliefs and squares denote possible actions. It searches all possible trajectories to find the optimal one (noted as the red path). Then the robot will execute the first action (noted as the pink square) with the highest expected cumulative reward.}
 \label{fig:treesearch}
 \end{figure}

Intuitively, our POMDP planner evaluates the trade-off between gathering more information and directly retrieving the target. 
This setting is similar to the Tiger problem~\citep{kaelbling1998planning}.
Therefore, we utilize the policy tree search introduced by \cite{kaelbling1998planning}) as the POMDP planner.

As shown in \figref{fig:treesearch}, our POMDP planner takes the current belief $\belief_t$ as the input, and performs look-ahead search for the optimal action sequence $a^*$ that maximizes the cumulative reward:
\begin{align}
a^*=\arg\max_{a}E\left[\sum_{t}^{\infty} \rew(\state_t, \action_t)\right]
\end{align}
In our policy tree, each node $\belief_t$ represents a belief.
The parent node $\belief_t$ and child node $\belief_{t+1}$ are connected with an observation-action pair.
Since $\gaction$ results in a terminal state, observation-action pairs are all based on $\qaction$ during planning.
The maximum search depth is set to 3 to limit the number of questions the robot can ask.
By traversing all possible trajectories, the planner returns an optimal trajectory that maximizes the expected cumulative reward (denoted as the red path in \figref{fig:treesearch}).
The robot then executes the first action in the optimal trajectory (denoted as the pink square).
If the action is a grasp macro, the robot grasps the first object in the grasp sequence. If the action is to ask a question, the robot simply says the caption generated by \cnet.
After the action is performed, the robot transits into the next step where it receives a new observation, updates its belief, and performs the search again.

\section{Experimental Setup}



\subsection{Implementation Details}
We deploy~\invigorate on a Fetch robot under the framework of the Robot Operating System (ROS). All deep neural networks run on a single external NVIDIA Titan X GPU.
We use Intel Realsense D435 camera to capture RGB images for visual inputs and point cloud for grasping and Google Cloud APIs to translate human verbal instructions into texts as well as synthesize speech for generated questions.


\subsection{Benchmark} 

\begin{table*}[t] 
\caption{Comparison of overall performance.}
\label{table:compres}
\begin{center}
\begin{tabularx}{1\textwidth}{l@{\extracolsep{\fill}}@{\hspace*{40pt}}ccccccc}
\toprule
& \multicolumn{3}{c}{Success Rate ($\uparrow$)} & \multicolumn{3}{c}{Number of Questions ($\downarrow$)} & \multirow{2}{*}{Reward ($\uparrow$)}\\ 
\specialrule{0em}{1pt}{1pt}
\cline{2-4}   \cline{5-7} 
\specialrule{0em}{1pt}{1pt}
 & \randomtest & \hardtest & \overalltest  & \randomtest & \hardtest & \overalltest \\
\midrule
MAttNet+VMRN & 0.76 & 0.60 & 0.68 & - & - & - & -3.20\\
w/o interaction & 0.74 & 0.58 & 0.66 & - & - & - & -3.40\\
w/o pointing & 0.76 & 0.72 & 0.74 & 0.54 & 0.61 & 0.55 & -3.40\\
w/o history & 0.86 & 0.84 & 0.85 & 1.26 & 2.02 & 1.69 & -4.38\\
w/o visual history & 0.86 & 0.78 & 0.82 & 0.74 & 0.81 & 0.76 & -3.04 \\
w/o tree search & 0.82 & 0.72 & 0.77 & 0.53 & 0.62 & 0.57 & -3.16\\
\midrule
\bf INVIGORATE & 0.86 & 0.80 & 0.83 & 0.63 & 0.69 & 0.65 &  -2.78\\
\bottomrule
\end{tabularx}
\end{center}
\end{table*}

\begin{figure}
  \centering
  \begin{tabular}{c}
  \includegraphics[width=\columnwidth]{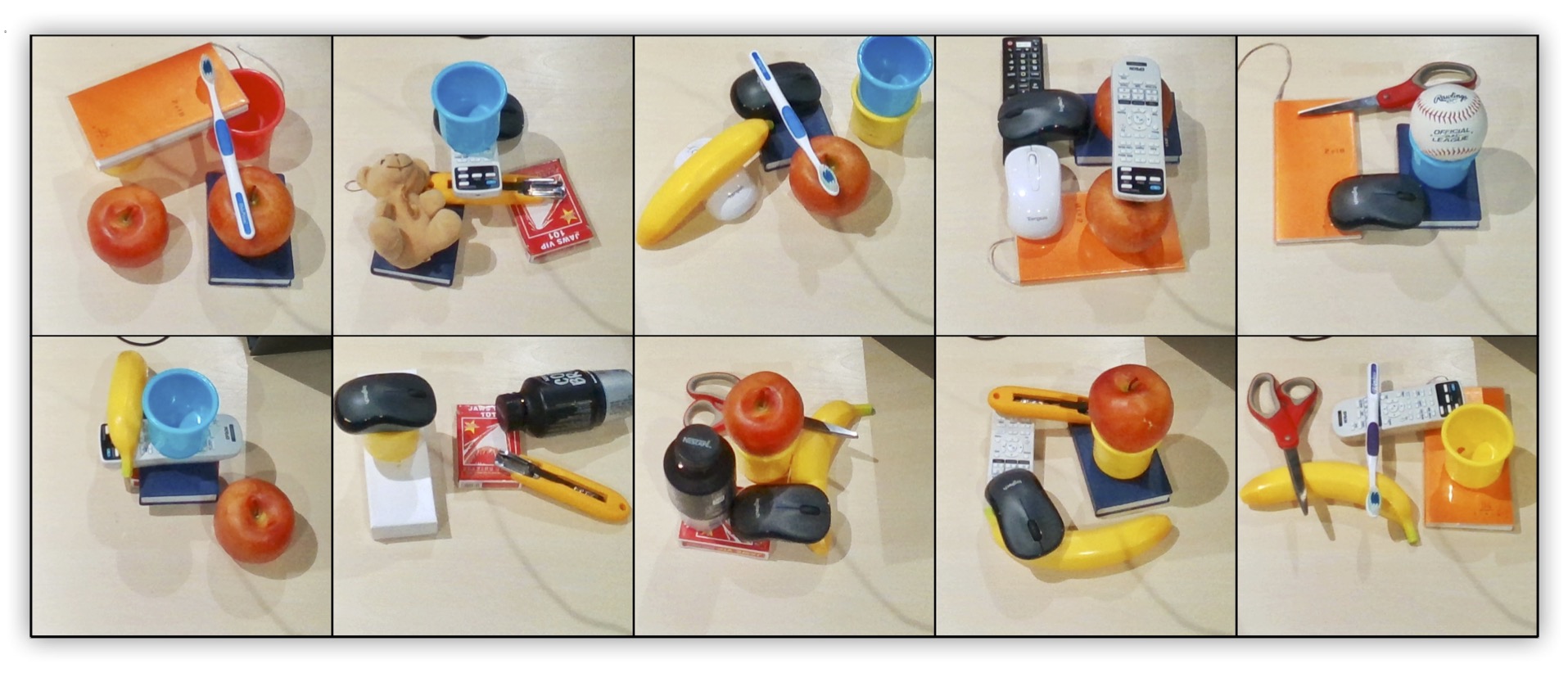}
  \end{tabular}
   \caption{Test dataset, consisting of 10 scenes in toal. The test dataset will be  available online.}
\label{fig:testexamples}
\end{figure}

To ensure fair comparisons between different variants of the system, we run all experiments on a test dataset consisting of 10 cluttered scenes shown in \figref{fig:testexamples}. We generate 100 test cases by recruiting 10 participants and asking them to select a target object and give a corresponding description for each scenario. 
For a comprehensive evaluation, we split test cases into two parts:
\begin{enumerate}
\item \textbf{\randomtest:} Targets are selected before the participants see the clutter but are described after the clutter is shown. Participants therefore do not know where the target object will be located at the time of target selection. 
\item \textbf{\hardtest:} Targets are selected by participants after they see the clutter. Participants exactly know which object is challenging for the robot to grasp.
\end{enumerate}
As we encourage participants to choose challenging targets, \textit{\hardtest} is generally harder than \textit{\randomtest}.


\subsection{Baseline}
~\invigorate combines data-driven learning and model-based planning. Given the success of deep learning, the natural tendency is to use it directly.
We build up the baseline method based purely on learned NN modules.
The baseline, called \textit{MAttNet+VMRN}, utilizes MAttNet \citep{yu2018mattnet} for visual grounding and VMRN \citep{zhang2019multi} for OBR and grasp detection.
It greedily follows the output of MAttNet to locate the most probable target while following the most likely OBRs to plan the grasp sequence. 

\subsection{Ablations}
~\invigorate POMDP consists mainly of three components: interaction, belief tracking, and policy tree search. 
In ablation studies, we aim to determine their respective contribution. 
Our ablation studies include:
\begin{itemize}
\item \textbf{w/o interaction: } the robot never asks questions but maintains the visual history to track the belief.
\bluetxt{
\item \textbf{w/o pointing: } the robot asks questions without pointing to the corresponding object. The question will be generated using the critiqued captioner.}
\item \textbf{w/o history: } the robot remembers neither visual nor QA history. The POMDP planner works on the belief estimated only by the current observation.
\item \textbf{w/o visual history: } the robot remembers QA history but not historical visual observations. The POMDP planner works on the belief estimated by the current visual observation and QA history.
\item \textbf{w/o tree search: } it utilizes a heuristic method instead of tree search. 
Concretely, we apply two-class K-Means to the expected rewards of all grasp macros to check if multiple grasp macros have similar expected rewards. 
If that is the case, the robot will ask a question. Otherwise, it will execute the grasp macro with the max reward.
\end{itemize}

\subsection{Procedures}

We conduct three experiments on the real robot using the collected dataset. 

The first experiment aims to compare the overall performance of~\invigorate against the baseline and conduct ablation studies. 
Each variant of the system receives the same initial image and expression from the dataset as input. It then computes an action for the robot to execute. 
{In each experimental scene, the experimenter is only required to describe one of the objects freely using its name, without any further detailed instructions to avoid possible interaction biases.}
During the process, if a question is asked, the experimenter will provide an answer (e.g., ``yes/no") according to whether the object being asked is the true target.
{Though the experimenter is allowed to give additional descriptions when being asked, we found that in our experiments they did not tend to do so.
Therefore, if there exist multiple ambiguous objects, the robot might ask several rounds of questions to disambiguate.}
Since grasp failures are not handled by any variant of the system and do not offer a meaningful comparison, if the robot fails to grasp an object, the experimenter would manually remove it.
We record the success rate of each variant. 
A test case is regarded as a success only if the robot retrieves the true target.
For ablation studies, we in addition record the \textit{Normalized Cumulative Reward} and \textit{Number of Questions} to give a comprehensive comparison.

The second experiment aims to compare~\invigorate's visual grounding performance against the SOTA method.
We run~\invigorate and ViLBERT side-by-side. 
No action is planned or executed by~\invigorate. 
The experimenter instead manually removes blocking objects sequentially to retrieve the final target.
In each step, we record the target probabilities estimated by both systems.
Since ViLBERT is trained with cross-entropy loss, we directly apply exponential on its output to get the target probabilities. 

\bluetxt{The third experiment aims to compare~\invigorate's captioning performance against baseline methods.
Concretely, we run the captioner in \invigorate to generate captions for each object in our collected dataset.
To compare the captioning performance, we first label the real-robot dataset manually with a detailed and distinct caption for each object bounding box.
Based on the labeled captions, we evaluate the quality of the generated captions from different captioners using several well-known metrics, including BLEU \citep{papineni2002bleu}, METEOR \citep{banerjee2005meteor}, ROUGH \citep{lin2004rouge}, CIDEr \citep{vedantam2015cider}, and SPICE \citep{anderson2016spice}.
Since our question generator is based on DenseCap \citep{johnson2016densecap} and UMD RefExp \citep{nagaraja2016modeling}, we mainly compare with these two baselines in this part.
}

\section{Experimental Results}

Our experiments  investigate four questions:
\begin{itemize}
    \item[Q1.] Does \invigorate perform well overall in interactive visual grounding and grasping tasks?
    \item[Q2.] What are the main contributors to \invigorate's performance?
    \item[Q3.] Does \invigorate perform well in visual grounding in clutter, a key component of the system? 
    \bluetxt{
    \item[Q4.]  Does critiqued question generation perform better than  SOTA
      captioners in clutter scenes?}
    \end{itemize}
For Q1, we compare the performance of~\invigorate with a pure deep-learning method without POMDP planning. Results show that~\invigorate outperforms the baseline substantially and achieves an overall 83\% success rate. 
For Q2, we conduct several ablation studies to evaluate various aspects of \invigorate. Results show that language interaction and observation histories boost overall success.
For Q3, we compare \invigorate with  ViLBERT~\citep{lu2019vilbert}, the current state-of-the-art visual grounding algorithm and show \invigorate consistently outperforms.
\bluetxt{For Q4, we compare the question generated using the critiqued captioner and the question directly generated from DenseCap or UMD RefExp.
Both the quantitative and qualitative results show that our critiqued question generation provide better performance.}

\subsection{Overall Performance }

\bluetxt{
In this section, we  address the first question: Does \invigorate perform well overall in interactive visual grounding and grasping?}

{~\tabref{table:compres} shows that \invigorate outperforms the baseline with an overall success rate of 83\% ($p < 0.01$ in t-test).} And in both \randomtest and \hardtest, \invigorate achieves higher success rates. On average, \invigorate asks 0.65 questions and spends 0.5 additional grasp steps per scenario. This shows~\invigorate achieves a higher success rate without a large number of redundant actions.

Furthermore, \invigorate's performance is more stable than the baseline. While the baseline \textit{MAttNet+VMRN} achieves an average success rate of 76\% on \randomtest, its performance drops severely to 60\% when applied on the harder \hardtest. 
In contrast, the performance of~\invigorate only drops by 6\%. 
A closer look at the experiment result shows that the baseline's performance drop in \hardtest is mainly due to the increase in target detection failures. 
In fact, the baseline nearly fails in all cases where the target object is not visible or not detected at the beginning. 
In such cases, without a probabilistic estimate of the true underlying state, the baseline simply chooses the most likely target among visible objects and retrieves it for the user.
On the other hand,~\invigorate is able to reason that the target is not directly visible and would choose the clearing action to look for the target at the bottom.

\subsection{Ablation Studies}

\begin{table*}
\caption{Comparison of visual grounding performance.}
\label{table:groundres}
\begin{center}
\begin{tabularx}{1\textwidth}{l@{\hspace*{50pt}}c@{\hspace*{40pt}}c@{\hspace*{40pt}}c@{\hspace*{40pt}}c@{\hspace*{40pt}}c@{\hspace*{40pt}}c}
\toprule
 & \multicolumn{3}{c}{Mean Accuracy} & \multicolumn{3}{c}{Mean L1 Loss} \\
 \specialrule{0em}{1pt}{1pt}
 \cline{2-4} \cline{5-7}
  \specialrule{0em}{1pt}{1pt}
 & \randomtest & \hardtest & \overalltest &\randomtest & \hardtest & \overalltest \\
\midrule
ViLBERT & 0.855 & 0.817 & 0.831 & 0.084 & 0.108 & 0.099 \\
\bf INVIGORATE & \bf0.879 & \bf0.873 & \bf0.875 & \bf0.049 & \bf0.050 & \bf0.050\\
\bottomrule
\end{tabularx}
\end{center}
\end{table*}

\begin{table*}
\caption{Comparison of question generation performance.}
\label{table:cap-results}
\begin{center}
\begin{tabularx}{1\textwidth}{l@{\hspace*{10pt}}c@{\hspace*{13pt}}c@{\hspace*{13pt}}c@{\hspace*{13pt}}c@{\hspace*{13pt}}c@{\hspace*{13pt}}c@{\hspace*{13pt}}c@{\hspace*{13pt}}c@{\hspace*{13pt}}}
\toprule
 & BLEU-1 & BLEU-2 & BLEU-3 &BLEU-4 & METEOR & ROUGE & CIDEr & SPICE \\
\midrule
DenseCap + Class Name & 0.150 & 0.112 & 0.095 & 0.000 & 0.158 & 0.433 & 1.480 & 0.329 \\
UMD RefExp + Class Name & \bf 0.429 & 0.273 & 0.210 & 0.162 & 0.185 & \bf 0.569 & 2.056 & 0.377 \\
\bf INVIGORATE & 0.363 & \bf 0.285 & \bf 0.240 & \bf 0.193 & \bf 0.213 & 0.561 & \bf 2.456 & \bf 0.420 \\
\bottomrule
\end{tabularx}
\end{center}
\end{table*}

 
\bluetxt{We now answer the second question, which aims to identify the main contributors the performance of \invigorate.}

\tabref{table:compres} also shows the results of ablation studies. We found that interaction significantly improves the overall success rate ($p<0.01$ in t-test). \textit{w/o Interaction} suffers about 17\% success rate loss, mainly from grasping the wrong target. The information gathered from interaction greatly helps to obtain an accurate belief and prevents the robot from target failures.
\bluetxt{From the comparison between {\it w/o pointing} and \invigorate, it is also noticeable that the pointing action is important for disambiguation.
The pointing action can bridge the gap between the robot and the user to understand the same description, which is sometimes ambiguous.
On the other hand, from the comparison between {\it w/o pointing} and {\it w/o interaction}, we can conclude that interaction is helpful even without pointing actions.
}

{In addition, we conclude that history reduces the number of questions. 
Compared to~\invigorate, \textit{w/o History} and \textit{w/o Visual History} ask more questions (both with $p<0.01$ in t-test). }
In \textit{w/o Visual History}, the robot uses only the current observation to estimate the belief over the state which is less accurate. 
Therefore, it has to ask more questions to refine its belief.
Besides, \textit{w/o History} asks the most number of questions as the robot does not remember previous answers from the human. Though it achieves a high success rate, the system's behavior is annoying, resulting in a low cumulative reward. 

{For comparison between \invigorate and \textit{w/o  tree search}, we found that only 100 experiments do not show a statistically significant difference due to high variance. Therefore, we conducted 100 more experiments with the same procedure.
Compared to \textit{w/o  tree search},~\invigorate asks slightly more questions (with $p<0.01$) but leads to fewer failures (with $p<0.05$).  
Noteworthily, \invigorate shows a higher cumulative reward than \textit{w/o  tree search}.
In our experiments, we also observed that the behavior of \textit{w/o  tree search} is more aggressive, which means that it tends to be confident about its judgment without asking questions.
The intrinsic reason should lie in the two-class K-Means policy, which is quite close to the one-step planning and might be myopic.
Unfortunately, 200 experiments still fail to show some significant difference.
We will conduct more experiments in the future to explore the effects of the planner.}

\subsection{Visual Grounding}

In this section, we want to answer the following question: does \invigorate perform well in visual grounding in clutter,  a key component of the system?

We compare target probability L1 loss and mean average accuracy between~\invigorate and ViLBERT. Results are shown in \tabref{table:groundres}.
In order to calculate the accuracy of both systems, we treat visual grounding as a binary classification problem. The object is regarded as the target once its target probability is higher than a certain threshold.
The mean accuracy reported is computed by averaging accuracies computed on 9 different thresholds (0.1 to 0.9 with interval 0.1).

Our results show that the visual grounding performance of~\invigorate consistently outperforms ViLBERT in clutter. Despite its SOTA performance on visual grounding in uncluttered scenes, ViLBERT suffers from visual occlusions and language ambiguities in our test dataset and becomes inaccurate and unstable.
On the other hand,~\invigorate treats neural network's outputs as noisy observations. It learns an observation model and uses the Bayesian filter to constantly update its belief of the state across multiple steps. Our results confirm that such a principled approach for visual grounding exhibits more robust performance in clutter.

\subsection{Question Generation}

\bluetxt{
In this section, we compare the quality of question generation by various methods and provide examples to illustrate the advantage of our critiqued question generation method.
}

\bluetxt{
For baseline comparison, we use  DenseCap \citep{johnson2016densecap}
and UMD RefExp \citep{nagaraja2016modeling, shridhar2020ingress} to generate
referring expressions, as they form the basis of our question generator.
We fit the generated object referring expressions into a common template for question generation.
Our experiments indicate, however, that these two state-of-the-art methods
alone are not reliable in cluttered scenes and often
  generate wrong subject nouns for objects partially occluded.
We thus introduce two stronger baselines: {\it DenseCap + Class Name} and {\it
  UMD RefExp + Class Name}.
Since DenseCap usually generates reliable attributes, \eg, color, for an
interested object, we first parse the generated captions and extract the
adjectives. We then form the referring expression by concatenating the extracted attributes and the class name from \onet ({\it DenseCap + Class Name}).
Similarly, UMD RefExp is good at predicting relationships among
interested objects. We concatenate the generated relational phrase and the
class name from \onet ({\it UMD RefExp + Class Name}).
}

\bluetxt{
The results are reported in \tabref{table:cap-results}.
Following the common evaluation protocol of natural language processing, we
compare our critiqued question generation with the baselines on several
widely-used metrics for image captioning, including BLEU (word-level precision) \citep{papineni2002bleu}, ROUGE (word-level recall) \citep{lin2004rouge}, METEOR (word-level F1 score) \citep{banerjee2005meteor}, CIDEr \citep{vedantam2015cider}, and SPICE \citep{anderson2016spice}.
The two most recent metrics, CIDEr and SPICE, focus on the semantics of expressions  rather
than merely match individual words one by one. 
Also, BLEU-4 captures semantics better than  than BLEU-1, as BLEU-4 uses
multiple words to gain contextual information. 
Overall,  our critiqued question generation performs well 
at the semantic level, according to CIDEr and SPICE. 
At the word level, our performance is comparable to {\it UMD RefExp + Class
  Name} and much better than {\it DenseCap + Class Name}. 
}

\begin{figure}
 \center{\includegraphics[width=0.5\textwidth]{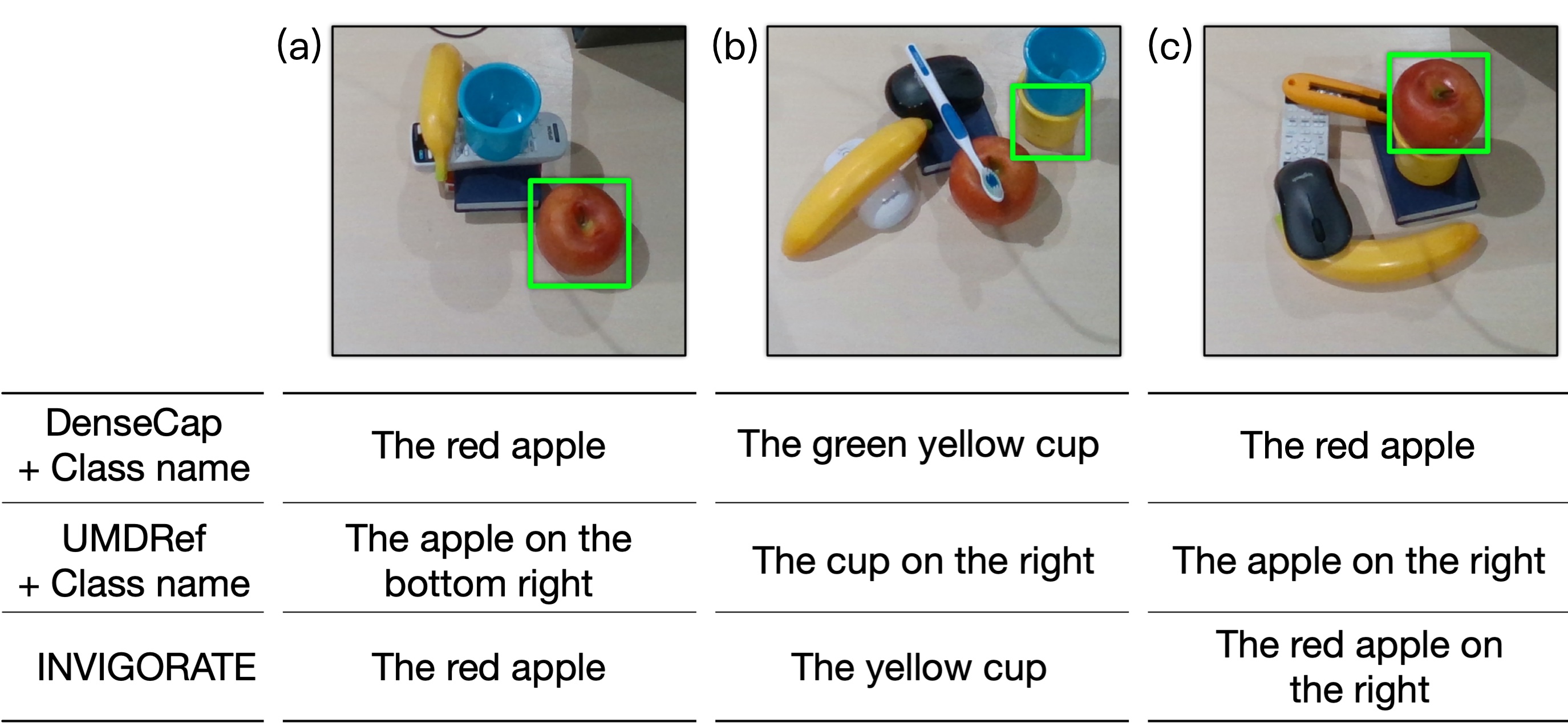}}
 \caption{Examples questions. Shown for each scene are the generated referring expressions, which are then inserted into a question template. For example, ``the red apple'' becomes ``Do you mean the red apple?''.}
 \label{fig:caption-result}
 \end{figure}
\bluetxt{
We also provide some examples to help understand the improved performance of critiqued question generation (\figref{fig:caption-result}).
Our method tends to favor the color as the primary attribute for object reference (\figref{fig:caption-result}\subfig{a--b}). This preference is consistent with that of humans, according to the  study of 
\cite{li2016spatial}, as our models are trained on human-labeled data and encode human preferences in the neural networks.
Further,  the critic filters out the incorrect ``hallucinated'' words (\figref{fig:caption-result}\subfig b).
However, the critic sometimes chooses a correct, but verbose expression 
(\figref{fig:caption-result}\subfig c), because it has not been trained to prefer concise expressions.
}

\subsection{Additional Examples}

\begin{figure*}[t]
  \captionsetup[subfloat]{farskip=2pt,captionskip=1pt}
  \centering
    \subfloat[]{\includegraphics[width=0.27\textwidth]{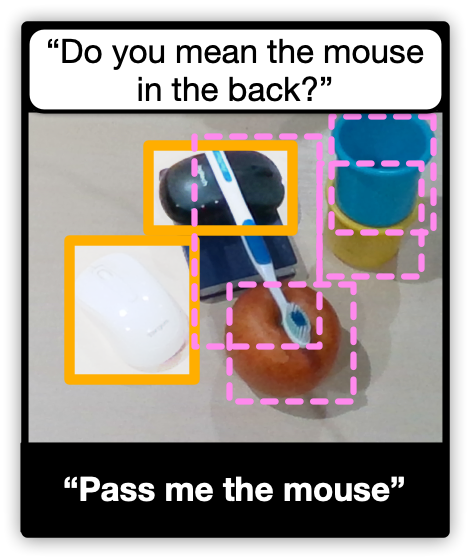}}
    \subfloat[]{\includegraphics[width=0.27\textwidth]{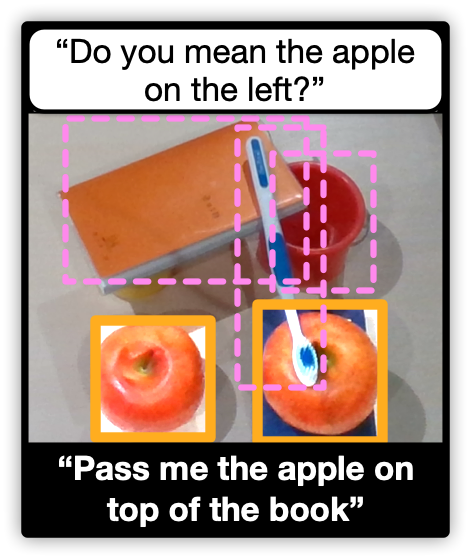}}
    \subfloat[]{\includegraphics[width=0.275\textwidth]{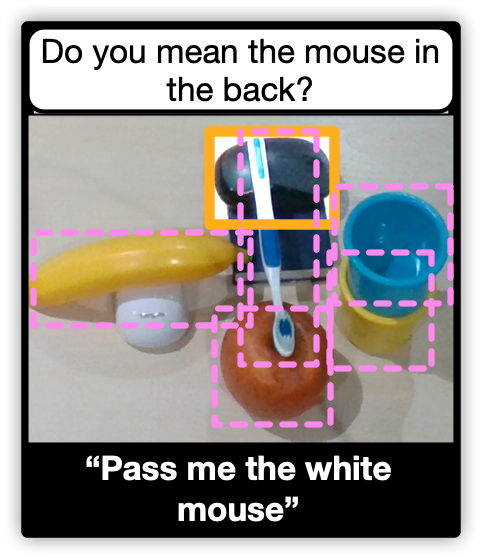}}
    
    \subfloat[]{\includegraphics[width=0.27\textwidth]{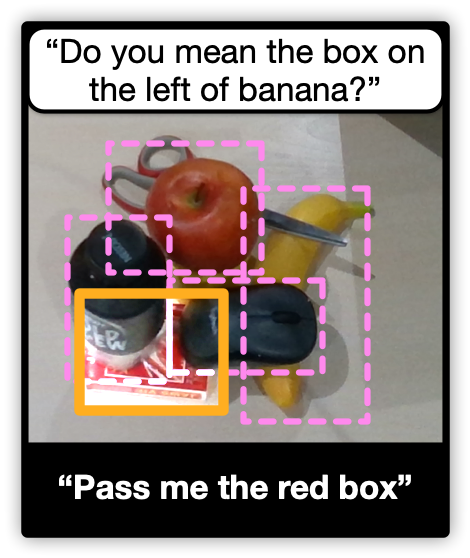}}
    \subfloat[]{\includegraphics[width=0.27\textwidth]{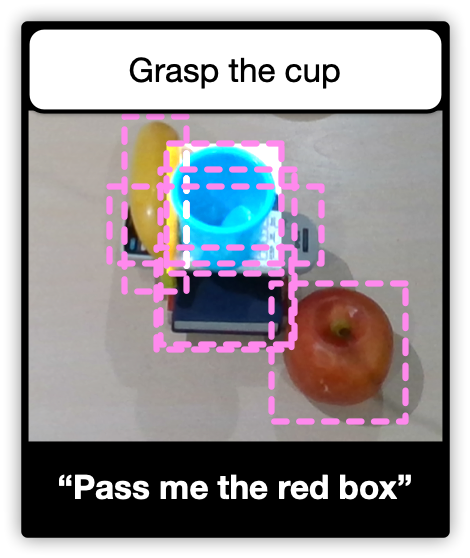}}
    \subfloat[]{\includegraphics[width=0.275\textwidth]{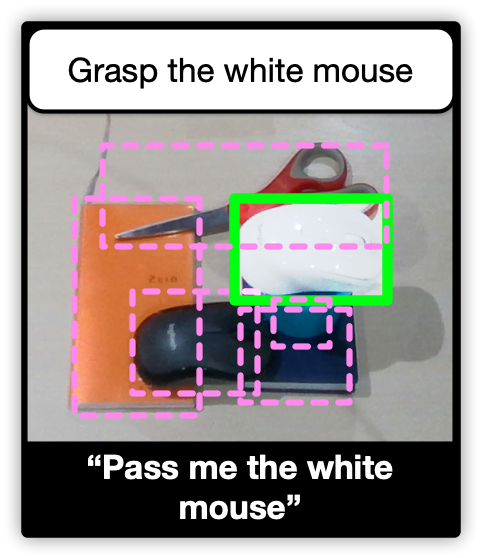}}
    
    \subfloat[]{\includegraphics[width=0.27\textwidth]{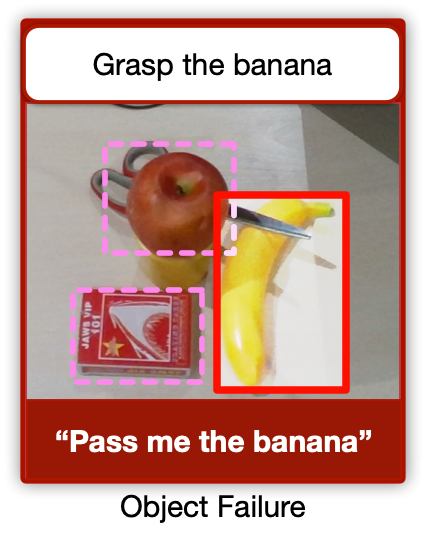}}
    \subfloat[]{\includegraphics[width=0.27\textwidth]{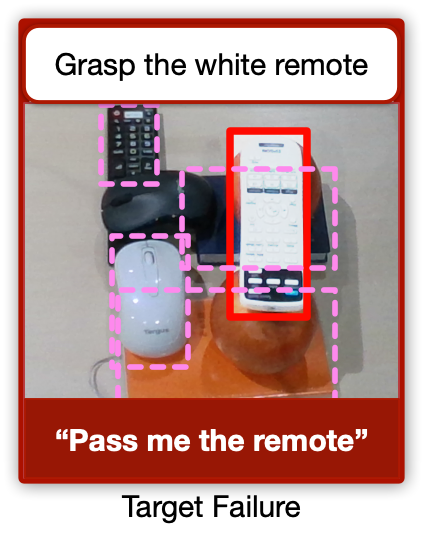}}
    \subfloat[]{\includegraphics[width=0.275\textwidth]{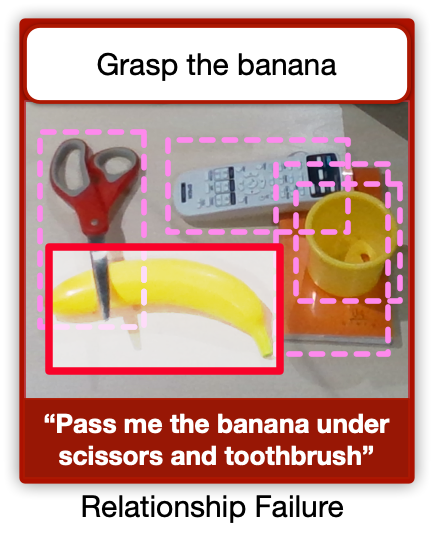}}
  \caption{Qualitative results. (\subfig a-\subfig f): some selected successful cases of \invigorate with different kinds of uncertainty and ambiguity. (\subfig g-\subfig i): some selected failures. The sentence on top is the action of the robot and the sentence at the bottom is the human's instruction. Best viewed in color.}
    \label{fig:test_examples}
\end{figure*}

\figref{fig:test_examples} shows examples of~\invigorate. 
In \figref{fig:test_examples}(a), the user gives an ambiguous expression. As there are two mice in the scene,~\invigorate asks a question to disambiguate. 
\figref{fig:test_examples}(b-d) show some complex scenarios where the target cannot be easily identified. In \figref{fig:test_examples}(b), the relational clue object ``book", which is the blue book lying under the right apple, is not detected. 
In \figref{fig:test_examples}(c), the black mouse is covered by a white toothbrush while the target white mouse is not detected. 
In \figref{fig:test_examples}(d), the red box is detected, but its visual features are not strong since it is occluded by the bottle and mouse on top. 
To tackle these difficulties,~\invigorate asks questions to query for more information.
\figref{fig:test_examples}(e) shows a case where the target is not detected,~\invigorate therefore removes the cup on top to look for the target at the bottom. 
\figref{fig:test_examples}(f) shows a simple case where the target is not occluded or obstructed,~\invigorate therefore directly grasps the target without asking any question. 

\figref{fig:test_examples}(g) shows that the object detector fails to detect the scissors that are blocking the true target banana.~\invigorate directly grasps the banana and thus violates the true OBR. In \figref{fig:test_examples}(h), the user gives an ambiguous expression ``remote" while the true target is the black remote in the back. Due to occlusion, the visual grounding module places a very low score on the true target,~\invigorate therefore grasps the white remote controller directly, believing that it is the only target in the scene. \figref{fig:test_examples}(i) shows a case of relationship detection failure.~\invigorate directly grasps the banana although it is blocked by the scissor, violating the true OBR.


\section{Conclusion}
\label{sec:conclusion}


\invigorate enables the robot to interact with human through the natural language and perform goal-directed object grasping in clutter.
It takes advantage of a POMDP model that integrates the learned neural network models for visual perception and language interaction. By integrating data-driven deep learning and model-based POMDP planning, \invigorate successfully tackles complex visual inputs and language interactions and achieves strong overall performance, despite the inevitable errors of the learned neural network models in perceptual and language processing. 



Many exciting challenges lie ahead. 
First, the neural network models for visual perception and language interaction are trained independently. It would be interesting to embed the \invigorate POMDP into the network and apply end-to-end training. 
Previous work demonstrates that such an end-to-end architecture may bring considerable performance improvement~\citep{karkus2017qmdp, tamar2016value}. 
Second, \invigorate cannot fully address the systematic errors from the deep neural network models for visual perception.
Calibration of the learned models~\citep{vaicenavicius2019evaluating}  can potentially  improve the uncertainty estimate for~\invigorate in the future. 
Finally, \invigorate assumes that the human would get annoyed if the robot asks more than 3 questions. It is reasonable simplification, but neglects the nuance of human-robot interaction. 
For humans, seamless interaction depends on conventions shaped  by shared experiences and culture.
For robots to  achieve the same, it may be necessary to model the human cognitive state and adapt information exchange accordingly \citep{goodrich2008human},  an interesting yet challenging direction for  future work.

\begin{acks}
This work has benefited greatly from discussions with Panpan Cai. It is supported in part by NSFC under grant No.91748208, No.62088102, No.61973246, Shaanxi Project under grant No.2018ZDCXLGY0607, the program of the Ministry of Education of China,
 Singapore A\(^*\)STAR under the National Robotics Program (Grant No. 192 25 00054), and  Singapore National Research Foundation under its AI Singapore Program (AISG Award No. AISG2-RP-2020-016).
\end{acks}

\bibliography{references}
\bibliographystyle{SageH}

\end{document}